\documentclass[11pt]{article}
\usepackage{coling2020}
\usepackage{times}
\usepackage{url}
\usepackage{latexsym}

\usepackage[final]{pdfpages}
\usepackage{flushend}

\usepackage{subfiles}

\usepackage{microtype}

\colingfinalcopy 

\title{MT-Adapted Datasheets for Datasets: \\Template and Repository\\}

\author{Marta R. Costa-juss\`a, Roger Creus, Oriol Domingo, Albert Dom\'inguez, \\ \textbf{Miquel Escobar, Cayetana L\'opez, Marina Garcia and Margarita Geleta} \\
  TALP Research Center \\
  Universitat Polit\`ecnica de Catalunya, Barcelona \\
  \texttt{marta.ruiz@upc.edu} \\}

\date{}

\begin{document}
\maketitle
\begin{abstract}
In this report we are taking the standardized model proposed by Gebru et al. \shortcite{gebru:2018} for documenting the popular machine translation datasets of the EuroParl \cite{koehn:2005} and News-Commentary \cite{barrault-etal-2019-findings}. Within this documentation process, we have adapted the original datasheet to the particular case of data consumers within the Machine Translation area.  We are also proposing a repository for collecting the adapted datasheets in this research area.
\end{abstract}

\section{Introduction}

Social biases are currently affecting the widely used natural language processing systems \cite{costajussa:2019}. While there are many proposed alternatives to mitigate this problem \cite{sun-etal-2019-mitigating}, there is still a long way to go \cite{gonen-goldberg-2019-lipstick-pig}. Research directions vary from debiasing algorithms \cite{bolukbasi:2016} to working directly towards fair or balanced datasets \cite{costajussa:2020}. While research community keeps active in these lines, there is an urgent need for transparency in our systems. As correctly addressed by the original work of Gebru et al. \shortcite{gebru:2018}, \textit{DataSheets for DataSets} proposes to create documentation for the datasets within the machine learning community to gain this transparency within research and in-production systems that are serving to different social purposes.

In our work, we want to use the existing datasheet template\footnote{https://www.overleaf.com/latex/templates/datasheet-for-dataset-template/ztkyvzddvxtd} and slightly adapt it to serve two main purposes: dataset usage in Machine Translation (MT) and dataset consumer-oriented. Our purpose is to motivate the community to work on these datasheets, independently of being dataset creators, in order to have proper documentation of the datasets that we are currently using. In fact, this report is the initiative for an open repository which aims at collecting the datasheets for MT datasets and can be accessed in here\footnote{https://mtdatasheets.cs.upc.edu}.

The rest of the report is organised as follows. The next section reports how we have modified the datasheet template
by both excluding and adding questions. Then, Section \ref{sec:repository} describes the repository to collect the datasheets which is open to contributions for documenting MT datasets. Finally, section \ref{sec:finalwords} reports some final words. Appendices report the datasheet for EuroParl \cite{koehn:2005} and News-Commentary \cite{barrault-etal-2019-findings}, two of the most popular MT datasets.

\section{DataSheet for DataSets: Adaptations}
\label{sec:excluded}

This section reports the main modifications done to the datasheet proposed by Gebru et al. \shortcite{gebru:2018} targeting MT datasets consumers. While we want to perform the minimum changes to the original datasheet template, we have two main purposes to perform this adaptation.

First, MT is clearly reporting biases \cite{prates}. While there are some solutions proposed from the algorithmic point of view \cite{Font:2019} and ways to properly evaluate the bias \cite{gabriel:2019}, there is no completed datasheet for any MT dataset since the first datasheet from Gebru et al \cite{gebru:2018} appeared. 

Second, MT is an already well-established area of research with a lot of existing resources that are not documented at all. For this, we want to adapt the datasheet more to consumers than to creators. 

At the end of the day, the final purpose of our work is to motivate the community to create this documentation, which is easier to do if the template targets MT (not artificial intelligence in general) and consumers (not to both consumers and creators) in particular. Finally, we motivate the community by collecting this documentation in an open repository (see Section \ref{sec:repository}). 

In the remaining of this section, we first report the excluded questions and second, we report the added questions. We report the modifications ordered by the inner datasheets sections which are: motivation; composition; collection process; data preprocessing/cleaning/labelling; distribution and maintenance. One must be aware that beyond the modifications reported in this section, the final datasheet, which is freely available in overleaf\footnote{https://www.overleaf.com/latex/templates/mt-adapted-datasheet-for-datasets-template/vjwbzfkpxthq}, also includes rewritten questions. 

\subsection{Excluded Questions}

\paragraph{Motivation}
We have included all questions.
\paragraph{Composition}
\begin{itemize}
    \item Are relationships between individual instances made explicit?
\end{itemize}
We have not included this question because  we are orienting the datasheets to data consumers. Then the person filling out the datasheet (hereinafter, expert), as data consumer and not creator, could only answer “yes” or “no” by observing its content, so it would not be informative enough.
\begin{itemize}
    \item Does the dataset identify any subpopulation (e.g. age, gender)?
    \item Is it possible to identify individuals (i.e. one or more natural persons) either directly or indirectly from the dataset?
    \item Does the dataset contain data that might be considered sensitive in any way (e.g. data that reveals racial or ethnic origins, sexual orientations, religious beliefs, political opinions or union memberships or locations; financial or health data biometric or genetic data, forms of government identification, such as social security numbers; criminal history)?
\end{itemize}
We have combined the last three questions into one. Since it would be
rather unlikely that the expert would have a deep
knowledge on these questions, we have only invited the expert to explain as
much details as one may know on the question.

\paragraph{Collection Process} 
\begin{itemize}
    \item How was the data associated with each instance acquired?
        \item What mechanisms or procedures were used to collect the data?
\end{itemize} 
Again, it is rather unlikely that the expert will have this
information. Therefore we have excluded this question but, again, invited the expert to explain it when possible.
\begin{itemize}
\item Who was involved in the data collection process and how were they
compensated?
\item Over what time frame was the data collected?
\end{itemize}
This information is rarely specified when publishing a dataset. Then, because
of the same reasons than before, we have decided to exclude them.

\paragraph{Preprocessing/Cleaning/Labelling}
We have excluded all the questions on this section because the expert will,
most likely, not know anything on this topic. The only chance that someone
who did not create the dataset could be aware of this level of detail, would
be if the creator had previously released information about it. That is why we
have invited the expert to specify any information available but have not
explicitly included the questions.

\paragraph{Use}
\begin{itemize}
    \item Is there anything about the composition of the dataset or the way it was
collected and preprocessed/cleaned/labeled that might im- pact future uses?
\end{itemize}
As we have discussed earlier, the expert will rarely
have much information on the collection and preprocessing stages and
therefore will most likely be unable to answer this.

\paragraph{Distribution}
We have excluded all questions under this topic but not the one referring to any
intelectual property licenses linked to the dataset either the one
referring to the dataset’s first releasing date.
Again, we have invited the expert to specify any other information on the
topic.

\paragraph{Maintenance}
\begin{itemize}
    \item Is there someone supporting/hosting/maintaining the dataset?
    \item If the dataset relates to people, are there applicable limits on the retention of
the data associated with the instances?
\end{itemize}
We have not explicitly included these questions but have invited the expert to
give any available information on it.

\begin{itemize}
    \item Will older versions of the dataset continue to be supported/hosted/maintained?
    \item If others want to extend/augment/build on/contribute to the dataset, is there a
mechanism for them to do so?
\end{itemize}
These questions, in the original datasheet form, are possibly intended to get some sort of
responsibility from the expert regarding the expiration of the data. It is
obvious that this is not the responsibility of the expert, that is why we have excluded these questions. However,
as always, we invite the expert to give any information on the topic if this was
publicly available.

\subsection{Added Questions}
\label{sec:added}

In the following, we list and explain several questions that we consider relevant for an MT-consumer datasheet.

\paragraph{Motivation}

We believe it is important to make the expert explicitly consider the data that one intends to use and, furthermore, study the ethics of it. Likewise, we want the expert to be aware of the legibility of the data, helping to evaluate whether it is representative or not for the task to be performed. We propose the following somehow subjective question.

\begin{itemize}
    \item Could any of these uses, or their results, interfere with human will or communicate a false reality?
\end{itemize}

We encourage the expert to study the precise motivation for creating the dataset.

\begin{itemize}
    \item What is the antiquity of the file? Provide, please, the current date.
\end{itemize}

Although we will later focus on the collection process we believe it is important to state, right at the begining, what sort of benefits, have the dataset creators got. 

\begin{itemize}
    \item Has there been any monetary profit from the creation of this dataset?
\end{itemize}

\paragraph{Composition}

We propose a high modification of the question \textit{What data does each instance consist of?}. Our proposed question pretends to broaden the scope of the original one and report whether the data is of a single type or of different ones, and also state what types are there. 

\begin{itemize}
    \item Are there multiple types of instances or is there just onetype?  Please  specify  the  type(s),  e.g.  Raw  data,  prepro-cessed, continuous, discrete
\end{itemize}

We have added a question that refers to which format is the data provided in.

\begin{itemize}
    \item What is the format of the data? e.g. .json, .xml, .csv
\end{itemize}

It is known that MT datasets do not include rare or uncommon languages as many times as
worldwide spoken languages, such as English. Not only are these datasets unbalanced in terms of languages, but they are also biased in terms of vocabulary. Hence, we consider the following questions to be relevant in the MT area:

\begin{itemize}
    \item Does the dataset cover included languages equitatively?
\item Is there any evidence that the data may be somehow biased (e.g. towards gender, ethics)?

\end{itemize}
It would also be good to consider the type of text contained in the data. This can also indirectly
bias MT algorithms, for example in terms of representation of some used vocabulary (such as contractions) or the presence of idioms.

\begin{itemize}
    \item Is the data made up of formal text, informal text or both equitably?
\end{itemize}
Furthermore, maybe the data is not completely made up of fully correct language and there is the
possibility of it containing, for example, extremely informal language. This can be desirable for some applications, but non-desirable for
others. In the case of MT, it could be useful to try and make the MT user avoid having to write
perfectly in order to get quality translations in an environment where the MT user is not used to
doing so.

\begin{itemize}
    \item Does the data contain incorrect language expressions on purpose? Does it contain slang terms? If that’s the case, please provide which instances of the data correspond to these.

\end{itemize}

Another common behaviour in MT is the inheritance of bias, which is covered in the following question.

\begin{itemize}
    \item  Is there any verification that guarantees there is not institutionalization  of  unfair  biases?  Both  regarding  the  dataset itself and the potential algorithms that could use it.
\end{itemize}

The preprocessing process will probably be the reason why a dataset is
a subset of another one, so it can be helpful to ask it explicitly so
that the expert thinks about it. It is different to have a whole dataset
and take only some instances to create another one (that is, a subset) than 
to have a dataset and that during the preprocess stage some data gets collapsed/deleted/transformed creating a dataset which is not the original one. 
\begin{itemize}
    \item  We have modified the following question \textit{Does the dataset contain all possible instances or is it just a sample of a larger set?} to Does the dataset contain all possible instances or is it just a sample of a larger set? Is the dataset different than an original one due to the preprocessing process? In case this dataset is a subset of another one, is the original dataset available?

\end{itemize}

We add a question referring to whether there has been any data
augmentation process performed at all.

\begin{itemize}
    \item  Is there synthetic data in the dataset? If so, in what percentage?
\end{itemize}

\paragraph{Collection Process}

The following two questions motivate the expert to get informed about the collection process and its legibility and at the same time will give an idea of the compromise of the dataset creators with both general law and individual rights.
\begin{itemize}
    \item Are there any guarantees that prove that the acquisition of the data did not violate any law or anyone's rights?
    \item Are there any guarantees that prove that the data is reliable?
\end{itemize}

We find that the question \textit{Does the dataset relate to people}, as well as the rest of questions in this subsection, will be already answered in the composition section, for this reason, we have modified this one so that widens the amount of information on the topic:

\begin{itemize}
    \item Did  the  collection  process  involve  the  participation  of individual people? ( e.g. Was the data collected from people directly?  Did  all  the  involved  parts  give  their  explicit  consent? Is there any mechanism available to revoke this consent in the future, if desired?)
\end{itemize}

The majority of data sources that are used in MT tasks, and more generally speaking in Natural Language Processing (NLP) tasks, consist of large corpora of plain text. We find it relevant to
carefully think if any variant on the source of the text (e.g. newspapers, speech transcriptions, novels) would give the same dataset as a result. Logically, as one gets more and more samples, the variances of the corpora extracted from different sources should tend to be smaller, but as an example, for the same politician, we may get different descriptive adjectives depending on the newspaper. Thus, we encourage dataset consumers to carefully think about these biased sources and collect representative datasets for the system they are targetting to build. 

Motivated by the previous statement, it is possible that a corpus is composed of data coming from different sources. In case the answer to the previous question is negative, then it is logical to ask the following questions.

\begin{itemize}
    \item  If the same content was to be extracted from a different source, would it be similar?
    \item Does the data come from a single source or is it the result of a combination of data
coming from different sources? In case it is a compilation of different sources, please
provide a link to those.
\end{itemize}

It may be relevant to point out where the collection process was performed (i.e. if the data was collected by a private company in some particular country).

\begin{itemize}
    \item Where was the data collected at? (Please include as much detail; i.e. entity, country, city, community, etc.) in order to include the where aspect.
\end{itemize}

\paragraph{Distribution}

MT datasets can be obtained/downloaded from different sources. These different sources are usually conventions, contents or events of such a type where the dataset has been used and published. However, we find it important to report where did the expert get the dataset from. 

\begin{itemize}
    \item Please specify the source where you got the dataset from.
\end{itemize}

We have already mentioned about the antiquity of the file. But, we believe it is useful, and necessary, to know not only the antiquity of the file in particular but the actual date of the first release of the dataset.
\begin{itemize}
    \item When was the dataset first released?
\end{itemize}

As dataset consumers/collectors we should care about widely sharing our datasets. Hence, we encourage to investigate if there is information about regions/countries where the dataset is not available or cannot be used.

\begin{itemize}
    \item Are there any restrictions regarding the distribution and/or usage of the data in any particular geographic regions?
\end{itemize}

\paragraph{Maintenance}

It is well known that all languages evolve over time. Recent sources of plain text may use vocabulary
following different distributions (e.g. trending topics, new vocabulary, and many other variants). It is
important to think about the possibility of the implementation of an automatic process that takes
care of these modifications.

\begin{itemize}
    \item {Are there any lifelong learning updates such as vocabulary enrichment automatically developed?}
\end{itemize}

It may be relevant to be able to access information on previous
versions and therefore it may be very handy to have it referenced if it exists.

\begin{itemize}
    \item Is there any log about the changes performed in
the dataset available?
\end{itemize}

There may be many different types of updates on the data that could be reported on a log file. For example, one may explicitly consider an update the removal of data when this has become irrelevant for the dataset usage. 
 
\begin{itemize}
    \item We have changed the question \textit{Is there any verified information on whether the
dataset will be updated in any form in the future?} to Is there any verified
information on whether the dataset will be updated in any form in the
future? Is someone in charge of checking if any of the data has become
irrelevant throughout the time? If so, will it be removed or labeled
somehow? 
\end{itemize}

MT and NLP in general are open source environments. Then it is very important to know whether the dataset is open for contributions and whether these may be done by anyone or only by a set of people. While this is somehow covered under the license the data is shared, we want to specifically state the following question to gain more insights on who can contribute to the dataset.

\begin{itemize}
    \item Specify any limitations to contributing to the dataset. 
\end{itemize}

Finally, it is clear that by the end of this section a deep study on the dataset maintenance will have been performed. However, there is another type of \textit{maintenance} that should also be check for: whether the dataset is or will still be legal throughout time. We have then added one last question referring to this:
\begin{itemize}
    \item Could changes to current legislation end the right-of-use of the dataset?
\end{itemize}

\section{Repository}
\label{sec:repository}

\begin{figure}[h]
\centering
\includegraphics[width=.5\textwidth]{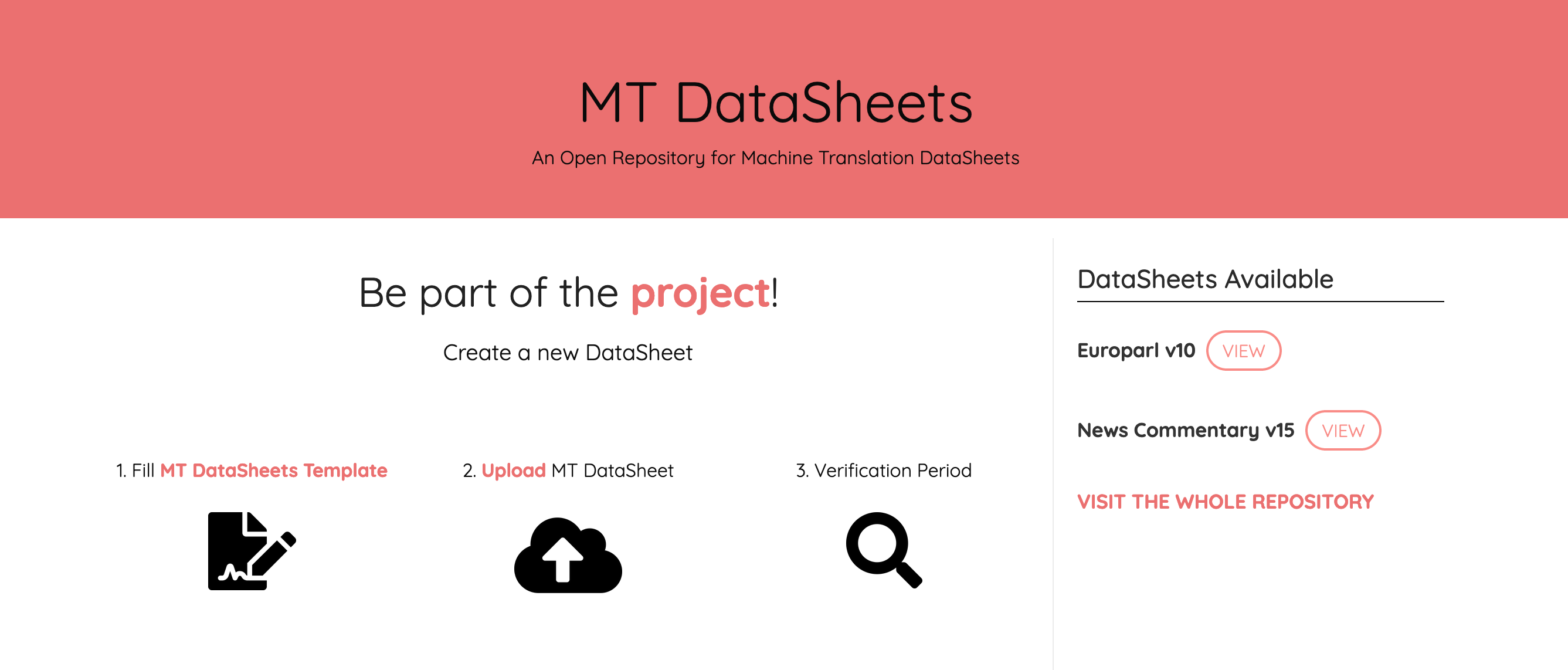}
\caption{Repository for the MT dataheets}
\label{fig:repository}
\end{figure}

The repository is made available in the following webpage: {\tt https://mtdatasheets.cs.upc.edu/}
The community can access and contribute to create new datasheets. Any uploaded datasheet undergoes a manual check-in procedure before finally made available in the webpage. Figure \ref{fig:repository} shows a snapshot of this repository.

\section{Conclusions}
\label{sec:finalwords}

This work pretends to push transparency in the specific area of Machine Translation by creating a repository of dataset documentation used in the area. This documentation slightly modifies the previously general datasheets \cite{gebru:2018} creating a new template available to the community in overleaf. 
 This paper describes the motivations to the modifications performed in the datasheet by discussing the deleted and added questions. Also, the appendix of this paper includes the first datasheets in the area for the popular corpus of EuroParl (v10)\footnote{http://www.statmt.org/europarl/v10/} and News-Commentary (v15)\footnote{http://data.statmt.org/news-commentary/v15/}, which are two of the corpus used in the WMT International Evaluation \cite{barrault-etal-2019-findings} as well as in many research papers. 

\section*{Acknowledgements}

We want to especially thank Barry Haddow for his feedback on the project; Christine Raouf Basta for her comments on the paper; our colleagues from Data Science Engineering for enriching discussions in the final datasheet and the members of RDLAB at UPC for their support with the repository. This work is supported in part by the Spanish Ministerio de Econom\'ia y Competitividad, the European Regional  Development  Fund  and  the  Agencia  Estatal  de  Investigaci\'on,  through  the  postdoctoral  senior grant Ram\'on y Cajal, the contract TEC2015-69266-P (MINECO/FEDER,EU) and the contract PCIN-2017-079 (AEI/MINECO).

\bibliography{emnlp2020}
\bibliographystyle{coling}

\appendix




\section*{Supplementary material}

# Europarl v10 Dataset Datasheet
## A collection of parallel corpora

### Disclaimer

This Datasheet has been inspired by [1] and modified as proposed by [2] and it is not filled out by the dataset creator. Therefore it is strongly recommended to only make use of this if the creator has not filled in a proper datasheet or to use it in combination. It is required that writers indicate their personal and contact data as well as the date this datasheet was last reviewed hereunder. Please, also remember to change the datasheet title to the name of the datatset in question.

This datasheet has been filled out by Marta R. Costa-jussà, Roger Creus, Oriol Domingo, Albert Domínguez, Miquel Escobar, Cayetana López, Marina Garcia and Margarita Geleta, from Universitat Politècnica de Catalunya (UPC) in Barcelona, which can be contacted at `marta.ruiz@upc.edu`.

This datasheet was last reviewed on May 25th 2020. Authors want to specially thank Barry Haddow for his feedback on this datasheet.

## I. Motivation

### A. Who created the dataset(e.g., which team, research group) and on behalf of which entity (e.g. company, institution, organization)?

The Europarl parallel corpus was created by a group of researchers led by Philipp Koehn at the University of Edinburgh [3].

### B. Did they fund it themselves? If there is an associated grant, please provide the name of the grantor and the grant name and number.

The construction of the European Parliament Proceedings Parallel corpus was supported by the EuroMatrix project and later by the following EuroMatrixPlus project funded by the European Commission (7th Framework Programme). It was also funded by the Mosescore[1] project which "encourages the development and usage of open source machine translation tools" and which is also supported by the European Commission Grant Number 288487 under the 7th Framework Programme.

---

[1] Find more about the Mosescore project at http://www.statmt.org/mosescore/ and https://github.com/moses-smt/mosesdecoder

### C. For what purpose was the data set created? Was there a specific task in mind? If so, please specify the result type ( e.g. unit ) to be expected.

EuroParl Parallel Corpus was created to provide a training data for corpus-based Machine Translation (MT) systems such as Statistical Machine Translation (SMT). However, nowadays there is barely no research on SMT, since almost everything is done with Neural Machine Trasnlation (NMT). And therefore we should be able to assume that this corpus is suggested to be used for these NMT tasks. Consequently the expected result would be the translation of the input, and therefore, text data.

### D. Could any of these uses, or their results, interfere with human will or communicate a false reality?

Although we are not aware of any specific communication on this topic, from our background knowledge we can be sure to warn that, due to the existence of bias in MT [4], the output text could be affected in such a way that it would not be portraying the actual essence of the translation. For example in the case of translating a sentence for which there is not explicit information on the subject's gender in one language while the translation must contain this gender information, due to the way gender is expressed in each of the languages[5].

Aside from this, one should also notice that the errors encountered in the dataset (e.g. misalignments, mixing of languages) could also report misleading results.

### E. What is the antiquity of the file? Provide, please, the current date.

The initial release of this corpus was back in 2005 and consisted of data up to 2001. It has since then been updated many times. The last update was on January the 17th 2020 and the data contained goes up to November 2011. At May 25th 2020.

### F. Has there been any monetary profit from the creation of this dataset?

This data was collected mainly to aid the authors' research in statisitcal MT and there is not evidence of getting any other profit from it.

### G. Any other comments?

The authors have shared that they "used the corpus to build 110 machine translation systems for all the possible

language pairs. The resulting systems and their performances demonstrate the different challenges for statistical MT for different language pairs"[3].

## II. COMPOSITION

### A. Is there any synthetic data in the dataset? If so, in what percentage?

No, there is not presence of synthetic data in this dataset.

### B. Are there multiple types of instances or is there just one type? Please specify the type(s), e.g. Raw data, preprocessed, symbolic.

At first the corpus was compound of raw data that had been crawled from the available content in the web [2]. However, after the preprocessing process this is converted to aligned sentences. Moreover, the dataset also contains metadata (i.e. the source, target, file ID, chapter ID, speaker ID, speaker name, language, and affiliation, which are strings and integers).

### C. What do the instances (of each type, if appropriate) that comprise the data set represent? (e.g. documents, photos, people, countries).

The text corpus is organised in documents, each entry indicates its content by the name (i.e. chapter ID indicates what document refers to) and then the corresponding text data in several source and target languages.

### D. How many instances (of each type, if appropriate) are there in total?

The number of instances (sentences) in parallel corpora highly depends on the languages that compose it. The range of instances varies from over few hundred thousands (Italian - Romanian) up to two million (French - English). As for this source [6], there are 21 languages available: Bulgarian, Czech, Danish, German, Lithuanian, English, Spanish, Estonian, Finnish, French, Hungarian, Italian, Lithuanian, Latvian, Dutch, Polish, Portuguese, Romanian, Slovak, Slovenian, Swedish.

In addition to the parallel corpus, there is also monolingual corpora available, and the range of instances varies from less than half a million for both Bulgarian and Romanian up to two and a half million for English.

### E. Does the dataset contain all possible instances or is it just a sample of a larger set? i.e. Is the dataset different than an original one due to the preprocessing process? In case this dataset is a subset of another one, is the original dataset available?

Although there may be more records on the proceedings of the European Parliament than the ones included in the corpus. It is self-contained in the sense that it is compound of all data that the creators retrieved to create the Europarl corpus.

### F. Is there a label or a target associated with each of the instances? If so, please provide a description.

Yes, in addition to the plain text in two languages, which is the core content of the dataset, there is some associated metada which includes the information of the structure of the proceedings and about the speaker [7].

### G. What is the format of the data? e.g. .json, .xml, .csv .

The data comes in a TSV (tab-separated values) file.

### H. Is any information missing from individual instances? If so, please provide a description, explaining why this information is missing (e.g. because it was unavailable). This does not include intentionally removed information, but might include, e.g. redacted text.

It has not been reported any issue of this kind.

### I. Are there any errors, sources of noise, or redundancies in the dataset? If so, please provide a description. Do not include missing information here.

It has been noticed that some special HTML entities and noisy characters have not been removed from the whole set of the Europarl corpus[3].

### J. Is there any verification that guarantees there is not institutionalization of unfair biases? Both regarding the dataset itself and the potential algorithms that could use it.

Since political discourses may portray aspects like personal opinions or generalizations, these could, either intentionally or not, generate biases on the data that could perpetuate throughout the translations. No mechanism to avoid biases has been used in the preprocessing process. Therefore, both because it is data from politics and because it is aimed to be used in MT could present bias in its content and also in the output it may produce.

### K. Are there recommended data splits, e.g. training, development/validation, testing? If so, please provide a description of these splits explaining the rationale behind them.

It is recommended to use data from the last quarter of 2000 as a test set, while the rest should be used as training data.

---

[2] http://www.europarl.eu.int/

[3] http://www.statmt.org/europarl/

*L. Is the dataset self-contained, or does it link to or otherwise rely on external resources? e.g., websites, tweets, other datasets. If it links to or relies on external resources, a) Are there any guarantees that they will exist, and remain constant over time? b) Are there official archival versions of the complete dataset? i.e. including the external resources as they existed at the time the dataset was created. c) Are there any restrictions (e.g. licenses, fees) associated with any of the external resources that might apply to a future user? Please provide descriptions of all external resources and any restrictions associated with them, as well as links or other access points, if appropriate.*

The dataset is self-contained in terms previously explained and therefore it is linked to the website of the European Parliament [4]. As it is compound of past records, there is an implicit guarantee that the data will remain constant. There are official reportings that can be found in the website of the European Parliament. The authors assure that they are not aware of the existence of any copyright restrictions of the material but encourage those that use the corpus to contact Philipp Koehn at pkoehn@inf.ed.ac.uk as indicated in the source website[5].

*M. Does the dataset contain data that might be considered confidential? e.g. data that is protected by legal privilege or by doctor patient confidentiality, data that includes the content of individuals non-public communications. If so, please provide a description.*

As data comes from European Parliament speeches, which are all available to the public, no confidential data is present.

*N. Does the dataset contain data that, if viewed directly, might be offensive, insulting, threatening, or might otherwise cause anxiety? If so, please describe why.*

It does not. However, several political points of views are present in the data, which may not be on par to someone's political views or may be considered extreme.

*O. Does the dataset relate to people? If so, please specify a) Whether the dataset identifies subpopulations or not. b) Whether the dataset identifies indivual people or not. c) Whether it contains information that could vulnerate any individuals or their rights. c) Any other verified information on the topic that can be provided.*

Because of the nature of the data, the corpus may reference many different people. On the other side, the data contained in the dataset, not only the corpus, does reference the individual speakers. However, it does not identify subpopulations and neither could vulnerate anyone's rights.

*P. Does the dataset cover included languages equally?*

[4]http://www.europarl.eu.int/
[5]http://www.statmt.org/europarl/

It is obvious that in parallel corpora both of the languages in question are equally covered.

However, as we have stated before, not all parallel corpora have the same size. As an example, while the French-English (FR-EN) corpus has over $1 \cdot 10^6$ instances (sentences), the Greek-English (EL-EN) corpus has a little less than $4.5 \cdot 10^5$ instances. As well as, if we compare monolingual corpora, the included languages are not covered equally. This is due to the lack of data in such a language in the crawling performed and therefore there is not a deliberate misbehaviour in any action regarding the dataset creation.

Because of this same reason, one should also notice that the content found in the different parallel corpora or monolingual corpora may not be the same.

*Q. Is there any evidence that the data may be somehow biased? i.e. towards gender, ethics, beliefs.*

Apparently, there is not, but as stated earlier, political-induced biases may be present. Aside from this inherited potential bias the errors found in the dataset could also cause biased results.

*R. Is the data made up of formal text, informal text or both equitably?*

Mostly formal since it is compound of speeches coming from proceedings at the European Parliament.

*S. Does the data contain incorrect language expressions on purpose? Does it contain slang terms? If that's the case, please provide which instances of the data correspond to these.*

One can assume it does not in the lines to which political speeches tend to be.

### III. COLLECTION PROCESS

*A. Where was the data collected at? Please include as much detail; i.e. country, city, community, entity and so on.*

As said before, the data contains speeches that took place at the European Parliament and therefore this data was generated either at Strasbourg, France or at Brussels, Belgium. The collection of the data, however, was originally performed at the University of Edinburgh [3].

*B. If the dataset is a sample from a larger set, what was the sampling strategy? i.e. deterministic, probabilistic with specific sampling probabilities.*

No specific sampling was performed as the dataset is not part from a larger set, as previously explained.

*C. Are there any guarantees that the acquisition of the data did not violate any law or anyone's rights?*

As the source data has been made public by the European Parliament itself, one should be able to assume that it is all in compliance with European laws.

*D. Are there any guarantees that prove the data is reliable?*

Despite being no guarantees, the fact that the data source is the European Parliament itself makes the data pretty reliable.

*E. Did the collection process involve the participation of individual people? If so, please report any information available regarding the following questions: Was the data collected from people directly? Did all the involved parts give their explicit consent? Is there any mechanism available to revoke this consent in the future, if desired?*

It does relate to people as every observation includes who is the speaker. The speaker name along with its affiliation is provided. The participants concern the gathering of the data according to European laws, as they are participating in public acts. As part of the public work of a civil servant, any potential mechanism that could exist for the speaker to revoke the data provided would be under the European Parliament's concern.

*F. Has an analysis of the potential impact of the dataset and its use on data subjects been conducted? i.e. a data protection impact analysis. If so, please provide a description of this analysis, including the outcomes, as well as a link or other access point to any supporting documentation.*

We have not found evidence of any analysis in this direction.

*G. Were any ethical review processes conducted?*

No ethical review processes were conducted, apparently.

*H. Does the data come from a single source or is it the result of a combination of data coming from different sources? In any case, please provide references.*

The data that makes up the corpus was extracted from the website of the European Parliament and then prepared for linguistic research. After sentence splitting and tokenization the sentences were aligned across languages with the help of an algorithm developed by Gale & Church[8].

*I. If the same content was to be collected from a different source, would it be similar?*

It definitely should, since it is official public data.

*J. Please specify any other information regarding the collection process. i.e. Who collected the data, whether they were compensated or not, what mechanisms were used. Please, only include if verified.*

As stated in [3], the acquisition of a parallel corpus for the use in MT typically takes the following steps[3]:
- Obtain the raw data (e.g., by crawling the web)
- Extract and map parallel chunks of text (document alignment)
- Break the text into sentences (sentence splitting)
- Map sentences in one language sentences in the other language (sentence alignment)
- Prepare the corpus for statistical MT systems (normalisation, tokenisation)

IV. PREPROCESSING/CLEANING/LABELLING

*A. Please specify any information regarding the preprocessing that you may know (e.g. the person who created the dataset has somehow explained it) or be able to find (e.g. there exists and informational site). Please, only include if verified. i.e. Was there any mechanism applied to obtain a neutral language? Were all instances preprocessed the same way?*

As mentioned, matching items were extracted and labeled with their corresponding document IDs [3]. As mentioned, sentence boundaries were automatically detected by using a preprocessing sentence boundaries. The data was sentence-aligned using the Gale & Church proposed algorithm [8].

The whole corpus has been preprocessed the same way.

V. USES

*A. Has the dataset been used already? If so, please provide a description.*

The dataset is free and it is available for commercial use and for research purposes. Surprisingly, bibliometric analyses show that it has hardly been used in translation studies, although this was the first purpose of its creators. Toolkits such as *EuroparlExtract* have been developed with this dataset [9].

Some works using this dataset include: "*Catalan-English Statistical Machine Translation without Parallel Corpus: Bridging through Spanish*" by Adrià de Gispert and José B. Mariño, and "*Bilingual Sentence Alignment of a Parallel Corpus by Using English as a Pivot Language*" by Josafá de Jesus Aguiar Pontes.

*B. Is there a repository that links to any or all papers or systems that use this dataset? If so, please provide a link or any other access point.*

There is not such specific repository. However, one can find the annual proceedings of WMT [6], with many papers reporting results based on this dataset.

*C. What (other) tasks could the dataset be used for? Please include your own intentions, if any.*

The dataset is intended for MT tasks. The original paper [3] puts emphasis on statistical MT tasks, but more

---
[6]http://www.statmt.org/wmt19/papers.html

generally this dataset is widely used for corpus-based MT, and nowadays is mostly used for Neural MT. Other NLP (Natural Language Processing) tasks such as language modeling, coreference resolution, name entity recognition can also benefit from this dataset. We are only completing this datasheet for the sake of providing it to the community. But we are not going to actually use the dataset for any task.

*D. Are there tasks for which the dataset should not be used? If so, please provide a description.*

There are no explicit tasks where the dataset should not be used.

*E. Any other comments? i.e. Do the collection or preprocessing processes impact future uses?*

There is minimal risk for harm since the data is already public under the reponsibility of the European Parliament.

## VI. Distribution

*A. Please specify the source where you got the dataset from.*

We have retrieved the corpus from the following sources: WMT [10] and Opus [6].

*B. When was the dataset first released?*

The initial release of this corpus was back in 2005.

*C. Are there any restrictions regarding the distribution and/or usage of this data in any particular geographic regions?*

The dataset was released publicly. It is freely available on the European Parliament website, and therefore it should be available in all regions.

*D. Is the dataset distributed under a copyright or other intellectual property (IP) license? And/or under applicable terms of use (ToU)? Please cite a verified source.*

Stated by the WMT source: "We are not aware of any copyright restrictions of the material.

## VII. Maintenance

*A. Is there any verified manner of contacting the creator of the dataset?*

Yes. All questions and comments can be sent to Philipp Koehn at pkoehn@inf.ed.ac.uk

*B. Specify any limitations there might be to contributing to the dataset. i.e. Can anyone contribute to it? Can someone do it at all?*

A number of researchers have adapted the Europarl corpus for specific purposes in order to further enhance its usefulness and to compensate for some of its drawbacks. For example, the organizers of WMT have been updating, removing irrelevant data and documenting all these changes from the dataset.

*C. Has any erratum been notified?*

No official specification has been made on the existence of errata in the Europarl corpus.

*D. Is there any verified information on whether the dataset will be updated in any form in the future? Is someone in charge of checking if any of the data has become irrelevant throughout time? If so, will it be removed or labeled somehow?*

Despite not being ensured, the wide use of the dataset and the fact that it has been updated more or less in a yearly basis during the last years, makes it highly probable that this data will be updated in the future. Although there is not a person in charge of ensuring the relevance of the data, researchers, usually, modify the dataset in order to make it more useful and meaningful for their projects.

*E. Is there any available log about the changes performed previously in the dataset?*

Yes, all the updates can be found at the source page: Scrolling down through the page one can find the versions http://www.statmt.org/europarl/archives.html#v1log, although not every version is fully available.

*F. Could changes to current legislation end the right-of-use of the dataset?*

Not in the foreseeable future, as it would imply that the European Parliament's data would not be public anymore.

*G. Are there any lifelong learning updates, such as vocabulary enrichment, automatically developed?*

Not that we are aware of.

# News-Commentary v15 Dataset Datasheet
A collection of parallel corpora

DISCLAIMER

This Datasheet has been inspired by [1] and modified as proposed by [2] and it is not filled out by the dataset creator. Therefore it is strongly recommended to only make use of this if the creator has not filled in a proper datasheet or to use it in combination. It is required that writers indicate their personal and contact data as well as the date this datasheet was last reviewed hereunder. Please, also remember to change the datasheet title to the name of the datatset in question.

This datasheet has been filled out by Marta R. Costa-jussà, Roger Creus, Oriol Domingo, Albert Domínguez, Miquel Escobar, Cayetana López, Marina Garcia and Margarita Geleta, from Universitat Politècnica de Catalunya (UPC) in Barcelona, which can be contact at marta.ruiz@upc.edu.

This datasheet was last reviewed on May 25th 2020. Authors want to specially thank Barry Haddow for his feedback on this datasheet.

## I. MOTIVATION

*A. Who created the dataset(e.g., which team, research group) and on behalf of which entity (e.g. company, institution, organization)?*

The parallel corpus of News Commentaries was provided by the University of Edinburgh, specifically the machine translation group [1].

*B. Did they fund it themselves? If there is an associated grant, please provide the name of the grantor and the grant name and number.*

The University of Edinburgh covered the expenses. It has never been attached to a specific grant.

*C. For what purpose was the data set created? Was there a specific task in mind? If so, please specify the result type ( e.g. unit ) to be expected.*

Different versions have been released since the first time this dataset was published. The original purpose was for the WMT International Evaluation Campaign [2]. Then, new versions have been released aiming to update the dataset by means of crawling the Project Syndicate website [3]. The main task from which this dataset was intended to be used was Machine Translation (MT).

*D. Could any of these uses, or their results, interfere with human will or communicate a false reality?*

It is known that the results in MT task can potentially communicate biased or unfair realities[3], [4]. The dataset is composed of political and economic commentaries from the Project Syndicate website, which is considered to be *The World's Opinion Page*. Thus, the source can inherit bias or unfairness from these articles related to the political and economic condition over the globe.

*E. What is the antiquity of the file? Provide, please, the current date.*

The released date of the latest versions, the 15th version, was in 2020[4]. At May 25th 2020.

*F. Has there been any monetary profit from the creation of this dataset?*

The dataset was released aiming to be useful for the Computational Linguistic research community in the field of MT but also in Natural Language Processing (NLP) in general. Nevertheless, information whether explicit monetary profit has been made or not could not be found.

## II. COMPOSITION

*A. Is there any synthetic data in the dataset? If so, in what percentage?*

Since this dataset is defined as a vanilla compilation of text, it does not contain synthetic data.

*B. Are there multiple types of instances or is there just one type? Please specify the type(s), e.g. Raw data, preprocessed, symbolic.*

All information found in the dataset consists of pieces of plain text for both parallel and monolingual corpora.

*C. What do the instances (of each type, if appropriate) that comprise the data set represent? (e.g. documents, photos, people, countries).*

---

[1]http://www.statmt.org
[2]http://www.statmt.org/wmt20/
[3]https://www.project-syndicate.org
[4]http://data.statmt.org/news-commentary/

They represent sentences which, in some cases, are compound of a single word. In the parallel corpus rows represent pairs of these mentioned types that have been aligned between two languages.

*D. How many instances (of each type, if appropriate) are there in total?*

The number of instances, sentences, in parallel corpora highly depends on the languages that compose it. The range of instances varies from around 1,500 sentences (Japanese - French) to almost 400,000 sentences (Spanish - English)[5]. As for this source, there is parallel corpus among 12 languages: Arabic, Czech, German, English, Spanish, French, Italian, Japanese, Dutch, Portuguese, Russian and Chinese.

For the monolingual corpora available, the range of instances varies from few thousands for Japanese to more than 600 thousand sentences for English[6].

*E. Does the dataset contain all possible instances or is it just a sample of a larger set? i.e. Is the dataset different than an original one due to the preprocessing process? In case this dataset is a subset of another one, is the original dataset available?*

It contains political and economic commentaries[7] crawled from the already mentioned Project Syndicate website. It is not a subset of any other published dataset, but there exist several versions which are based on different crawling executions. Therefore, although the dataset is not a subset of any other, it is a reduced crawling from the Project Syndicate website that only takes in to account the topics of Economics and Politics.

*F. Is there a label or a target associated with each of the instances? If so, please provide a description.*

Only information at the document level exists, except for that it is just a recompilation of plain text.

*G. What is the format of the data? e.g. .json, .xml, .csv .*

Files used in the monolingual corpus consist of .txt sources of plain text. The alignment units are saved in TMX files or also (more interpretable) in XML files, where the alignment between the positions of words in the sentences pairs is saved.

*H. Is any information missing from individual instances? If so, please provide a description, explaining why this information is missing (e.g. because it was unavailable). This does not include intentionally removed information, but might include, e.g. redacted text.*

There are some language pairs for which there is no parallel corpus at all, for example the Indonesian - Japanese[8].

[5] http://opus.nlpl.eu/
[6] http://data.statmt.org/news-commentary/v15/training-monolingual/
[7] http://www.casmacat.eu/corpus/news-commentary.html
[8] http://data.statmt.org/news-commentary/v15/training/?C=S;O=A

*I. Are there any errors, sources of noise, or redundancies in the dataset? If so, please provide a description. Do not include missing information here.*

There might be sentences that appear more than once in different news and contexts, although it is not considered a redundancy because it might be useful for some models or problems to understand the different applications of a specific sentence. With low probability, it can happen that in a given dataset may exist an instance, a sentence, in a different language than the expected.

*J. Is there any verification that guarantees there is not institutionalization of unfair biases? Both regarding the dataset itself and the potential algorithms that could use it.*

No, there is no verification. All the bias that is contained in the compilation of text from political and economic news will be present in the algorithms built on top of it.

*K. Are there recommended data splits, e.g. training, development/validation, testing? If so, please provide a description of these splits explaining the rationale behind them.*

There are development and test sets prepared for each WMT International Evaluation Campaign [5].

*L. Is the dataset self-contained, or does it link to or otherwise rely on external resources? e.g., websites, tweets, other datasets. If it links to or relies on external resources, a) Are there any guarantees that they will exist, and remain constant over time? b) Are there official archival versions of the complete dataset? i.e. including the external resources as they existed at the time the dataset was created. c) Are there any restrictions (e.g. licenses, fees) associated with any of the external resources that might apply to a future user? Please provide descriptions of all external resources and any restrictions associated with them, as well as links or other access points, if appropriate.*

It consists of political and economic news crawled from Project Syndicate. Once these are published they might never be modified again (maybe in case there was an error, which sounds unlikely since these are carefully reviewed before being published). There is no fee since it is and open source of news.

*M. Does the dataset contain data that might be considered confidential? e.g. data that is protected by legal privilege or by doctor patient confidentiality, data that includes the content of individuals non-public communications. If so, please provide a description.*

No it does not. As mentioned before they come from an open source of information.

*N. Does the dataset contain data that, if viewed directly, might be offensive, insulting, threatening, or might otherwise cause anxiety? If so, please describe why.*

It should not be the case, since the exact same information is shown to the public that reads the news through this source.

*O. Does the dataset relate to people? If so, please specify a) Whether the dataset identifies subpopulations or not. b) Whether the dataset identifies indivual people or not. c) Whether it contains information that could vulnerate any individuals or their rights. c) Any other verified information on the topic that can be provided.*

It might do but not in a targeted way, the people or individual referenced in every single article depends on the topic of it and on the relevant cause that has made it appear publicly.

*P. Does the dataset cover included languages equally?*

It is obvious that in parallel corpora both of the languages in question are equally covered.

However, as we have stated before, not all parallel corpora have the same size. As well as, if we compare monolingual corpora, the included languages are not covered equally. This is due to the lack of data in such a language in the crawling performed and therefore there is not a deliberate misbehaviour in any action regarding the dataset creation.

Because of this same reason, one should also notice that the content found in the different parallel corpora or monolingual corpora may not be the same.

*Q. Is there any evidence that the data may be somehow biased? i.e. towards gender, ethics, beliefs.*

It is widely criticised that newspapers give a subjective point of view in every new they publish, so there is the awareness that the possible bias in Project Syndicate relating politics and economics might be present.

*R. Is the data made up of formal text, informal text or both equitably?*

It is all formal text. It is written in the manner one would expect to see in an online newspaper.

*S. Does the data contain incorrect language expressions on purpose? Does it contain slang terms? If that's the case, please provide which instances of the data correspond to these.*

As mentioned before, it should contain appropriate language expressions due to the fact that is crawled text from an online newspaper.

### III. COLLECTION PROCESS

*A. Where was the data collected at? Please include as much detail; i.e. country, city, community, entity and so on.*

The text instances in dataset were crawled from the Project Syndicate website[9] which is compound of articles that talk about the economical, political, developing, sustainability-related, cultural or innovation situations of places around the world and, of course, these articles are written by many journalists, of all nationalities. Therefore the data may be considered to have been extracted from many places around the globe.

*B. If the dataset is a sample from a larger set, what was the sampling strategy? i.e. deterministic, probabilistic with specific sampling probabilities.*

It all comes from the crawling process of the same source.

*C. Are there any guarantees that the acquisition of the data did not violate any law or anyone's rights?*

The fact that all this data could be accessed at any given moment in the website it is extracted from implies that the acquisition is legal. If not used violating any of the website terms and conditions, then the whole process is also legal.

*D. Are there any guarantees that prove the data is reliable?*

Despite being no guarantees, the fact that the data source is the Project Syndicate makes the data pretty reliable.

*E. Did the collection process involve the participation of individual people? If so, please report any information available regarding the following questions: Was the data collected from people directly? Did all the involved parts give their explicit consent? Is there any mechanism available to revoke this consent in the future, if desired?*

There is the same security that exists in the newspaper polices. Any reclamation that might be sent to the newspaper after the crawling is done might affect the ethical purity of the data collection.

*F. Has an analysis of the potential impact of the dataset and its use on data subjects been conducted? i.e. a data protection impact analysis. If so, please provide a description of this analysis, including the outcomes, as well as a link or other access point to any supporting documentation.*

No, no analysis regarding this subject has been conducted.

*G. Were any ethical review processes conducted?*

No, there were not any ethical review processes conducted.

*H. Does the data come from a single source or is it the result of a combination of data coming from different sources? In any case, please provide references.*

The data comes from a single source, which is the Project Syndicate website, as stated here[10] however, the dataset was created by researchers at the University of Edinburgh.

*I. If the same content was to be collected from a different source, would it be similar?*

---

[9] https://www.project-syndicate.org

[10] http://www.casmacat.eu/corpus/news-commentary.html

Yes, the data would probably be very similar - given that newspapers' content tends to be about similar topics and thus, the used words are very similar. If the new source were to be different from a newspaper, there could be some significant differences when it comes to words use distribution and some specific translations. Even so, the law of large numbers implies that, if the source contains a high enough amount of data, the words distribution will tend to be very coincident between sources.

*J. Please specify any other information regarding the collection process. i.e. Who collected the data, whether they were compensated or not, what mechanisms were used. Please, only include if verified.*

The dataset release purpose was for the WMT International Evaluation Campaing. Nothing suggests that there has been any economical compensation. The mechanism consisted of a crawling process on the Project Syndicate website.

## IV. PREPROCESSING/CLEANING/LABELLING

*A. Please specify any information regarding the preprocessing that you may know (e.g. the person who created the dataset has somehow explained it) or be able to find (e.g. there exists and informational site). Please, only include if verified. i.e. Was there any mechanism applied to obtain a neutral language? Were all instances preprocessed the same way?*

The corpus is available in several formats: raw text files (not preprocessed) and sentence aligned files (with alignment preprocessing). No further information about preprocessing is given by the dataset creators.

The entire corpus has been preprocessed the same way.

## V. USES

*A. Has the dataset been used already? If so, please provide a description.*

This dataset has been used for all the editions of the WMT International Evaluation Campaings [5], which are annual events on MT. This is its main use, although it has also been used in many research papers.

The dataset has also been used in several projects, for example, the CASMACAT project (2012-2014) which built a translator's workbench to improve productivity, quality, and work practices in the translation industry. CASMACAT stands for *Cognitive Analysis and Statistical Methods for Advanced Computer Aided Translation* [11].

*B. Is there a repository that links to any or all papers or systems that use this dataset? If so, please provide a link or any other access point.*

The WMT event presents a findings paper for each edition [5], but beyond this, there is no specific repository that contains information of this kind.

*C. What (other) tasks could the dataset be used for? Please include your own intentions, if any.*

The dataset is intended for MT, even though it can also be used for other NLP tasks such as language modelling or language generation. We are only completing this datasheet for the sake of providing it to the community. But we are not going to actually use the dataset for any task.

*D. Are there tasks for which the dataset should not be used? If so, please provide a description.*

There are no explicit tasks where the dataset should not be used.

*E. Any other comments? i.e. Do the collection or preprocessing processes impact future uses?*

There is minimal risk for harm since it is publicly available data.

## VI. DISTRIBUTION

*A. Please specify the source where you got the dataset from.*

The source of the dataset are all the pages contained within the Project Syndicate[12] website.

*B. When was the dataset first released?*

The initial release of this corpus was back in 2007[13].

*C. Are there any restrictions regarding the distribution and/or usage of this data in any particular geographic regions?*

The dataset was released publicly and therefore it is available for all regions.

*D. Is the dataset distributed under a copyright or other intellectual property (IP) license? And/or under applicable terms of use (ToU)? Please cite a verified source.*

As stated by the authors, "No claims of intellectual property are made on the work of preparation of the corpus". See source.

---

[11] http://www.casmacat.eu/

[12] https://www.project-syndicate.org
[13] http://www.statmt.org/wmt07/shared-task.html

## VII. Maintenance

### A. Is there any verified manner of contacting the creator of the dataset?

There is no verified manner of contacting the creator as the latter is not known. The maintainer and responsible of the entire corpus can be contacting Barry Haddow (University of Edinburgh) at bhaddow@inf.ed.ac.uk.

### B. Specify any limitations there might be to contributing to the dataset. i.e. Can anyone contribute to it? Can someone do it at all?

There are not specific limitations, any contribution can be done by directly contacting Barry Haddow (as mentioned in the previous question). Mainly for notifying on errors.

### C. Has any erratum been notified?

No erratum has been notified. Data itself cannot contain errors, as it is simply the results of crawling an existing web, and it can be accessed with no errors by means of the provided files.

### D. Is there any verified information on whether the dataset will be updated in any form in the future? Is someone in charge of checking if any of the data has become irrelevant throughout time? If so, will it be removed or labeled somehow?

There is no verified information on this matter. Even so, the latest updates have been released recently which might indicate there is still room for more.

### E. Is there any available log about the changes performed previously in the dataset?

There is not any log of this type, in words of one of its creators (Barry Haddow), this is because the main changes are about enlarging the data.

### F. Could changes to current legislation end the right-of-use of the dataset?

In principle, there are low risks in this direction. But we could consider changes in the legislation of web crawling, since this could limit the methodology used for the data extraction. As well as changes in the restrictions of usage, specially the ones related to obtaining profit from the data, which could invalidate some tasks performed on it.

### G. Are there any lifelong learning updates, such as vocabulary enrichment, automatically developed?

No, there is no such mechanism. One of the main reasons being the fact that the dataset is a reflection of a website, and thus it simply contains what there is in it - with no external additions nor enrichment.



\end{document}